\documentclass{article}

\pdfoutput=1
\usepackage{authblk}
\usepackage[margin=1in]{geometry}
\usepackage[utf8]{inputenc}
\usepackage{graphicx}
\usepackage{amsmath}
\usepackage{amsfonts}
\usepackage{color, colortbl}
\usepackage[colorlinks = true,
            linkcolor = blue,
            urlcolor  = blue,
            citecolor = blue]{hyperref}

\usepackage[mathlines]{lineno}
\usepackage[round]{natbib}

\usepackage[T1]{fontenc}
\usepackage[utf8]{inputenc}
\usepackage{mathptmx}
\usepackage{multicol}

\usepackage[mathscr]{eucal}

\begin{document}

\begin{center}
    \huge
    Examining the causal structures of deep neural networks using information theory \\
    \vspace{0.2in}
    
    \normalsize
    Simon Mattsson$^{1,}$\footnote{Co-first author}, Eric J. Michaud$^{2,*}$, Erik Hoel$^{1,}$\footnote{Corresponding author: hoelerik@gmail.com}
    
    \vspace{0.2in}
    \small
    \textit{
        $^1$Allen Discovery Center, Tufts University, Medford, MA 02155, USA \\
        $^2$University of California, Berkeley, Berkeley, CA 94720, USA
    }
    \vspace{0.05in}
\end{center}


\begin{abstract}
  Deep Neural Networks (DNNs) are often examined at the level of their response to input, such as analyzing the mutual information between nodes and data sets. Yet DNNs can also be examined at the level of causation, exploring ``what does what'' within the layers of the network itself. Historically, analyzing the causal structure of DNNs has received less attention than understanding their responses to input. Yet definitionally, generalizability must be a function of a DNN's causal structure since it reflects how the DNN responds to unseen or even not-yet-defined future inputs. Here, we introduce a suite of metrics based on information theory to quantify and track changes in the causal structure of DNNs during training. Specifically, we introduce the effective information (\textit{EI}) of a feedforward DNN, which is the mutual information between layer input and output following a maximum-entropy perturbation. The \textit{EI} can be used to assess the degree of causal influence nodes and edges have over their downstream targets in each layer. We show that the \textit{EI} can be further decomposed in order to examine the sensitivity of a layer (measured by how well edges transmit perturbations) and the degeneracy of a layer (measured by how edge overlap interferes with transmission), along with estimates of the amount of integrated information of a layer. Together, these properties define where each layer lies in the ``causal plane'' which can be used to visualize how layer connectivity becomes more sensitive or degenerate over time, and how integration changes during training, revealing how the layer-by-layer causal structure differentiates. These results may help in understanding the generalization capabilities of DNNs and provide foundational tools for making DNNs both more generalizable and more explainable.
\end{abstract}

\section{\label{sec:introduction}Introduction}
Deep neural networks (DNNs) have shown state-of-the-art performance in varied domains such as speech synthesis \citep{wu2016merlin}, image recognition \citep{krizhevsky2012imagenet, xi2017capsule} and translation \citep{sutskever2014sequence}.
These immense advances have been due to the introduction of deep learning techniques \citep{lecun2015deep} to artificial neural networks, and the use of GPUs for high-speed computation \citep{raina2009large}. Yet the performance of DNNs remains mysterious in multiple ways. For instance, fundamental machine learning theory suggests that models with enough parameters to completely memorize large data sets of images should vastly overfit the training data and lead to poor generalization, especially in models that are not regularized \citep{zhang2016understanding}. However, in practice, deep neural networks have good generalization performance, even when not explicitly regularized \citep{neyshabur2017exploring}. While it is well known that artificial neural networks can approximate any given function \citep{hornik1989multilayer}, how the functions they arrive at generalize beyond their training data is less well understood.

One promising approach to explaining the generalization capability of DNNs is the information bottleneck approach \citep{tishby2000information}. The information bottleneck approach conceives of DNNs as optimizing the tradeoff between compression of input data into an internal representation and prediction of an output using this representation. Proponents of this approach analyze DNNs by their behavior in the ``information plane,'' composed of layer-to-input mutual information scores given a data set as input \citep{shwartz2017opening}. While looking for information bottlenecks has been a rich research program, larger networks are still plagued by information estimation issues \citep{wickstrom2019information}, and there have been errors in predictions or deviations for certain network topologies and activation functions \citep{saxe2019information}. More fundamentally, the information bottleneck approach is in its mathematical formulation data-dependent, that is, its mutual information scores vary with changes to input distributions. Yet generalizability is definitionally a function of performance across different data sets with different frequencies of inputs, or even unknown and not-yet-defined future data sets. Therefore to understand generalizability it is necessary to focus on what is invariant in DNNs across different data sets with different properties \citep{zhang2016understanding}. 

Examining what is independent across differing data sets means investigating the \textit{causal structure} of DNNs themselves. That is, uncovering the set of causal relationships (dependencies) between the nodes in the network using techniques from the field of causal analysis. Here we introduce a perturbational approach that uses information theory to track the causal influences within a DNN in a layer-by-layer manner. Specifically, we introduce the \textit{effective information} ($EI$), which captures the informativeness and therefore strength of a causal relationship. The $EI$ was originally introduced as a information-theoretic measure of the causal relationships between two subsets of a complex system \citep{tononi2003measuring}. $EI$ has already been shown to quantify the causal structure of Boolean networks \citep{hoel2013quantifying}, and also graphs, by measuring the amount of information contained in the dynamics of random walkers \citep{klein2020emergence}. Notably, $EI$ has mathematical similarities to the information bottleneck approach, although it is focused on causation and therefore differs in key ways. 

To measure the $EI$ between feedfoward layers of a DNN we bin the activation levels of nodes, inject independent and simultaneous white noise (maximum entropy) into a layer, then calculate the transmitted mutual information to the downstream targets. This captures the total amount of information in the causal structure of that layer-to-layer connectivity. Looking across network architectures, tasks, and activation functions, we observe that steep changes in the loss curve are reflected by steep changes in the $EI$. 

Additionally, $EI$ can be used to track how the causal structures of layers in DNNs change in characteristic ways during training. Specifically, we show how to track DNNs during training in the space of possible causal structures (the ``causal plane''), such as whether the connectivity becomes more informationally degenerate or more sensitive. This allows us to show how DNNs develop specific layer-by-layer causal structures as they are trained. We hypothesize that the differentiation of layer-by-layer causal structure may assist generalizability, as networks trained on simpler tasks show less differentiation than those trained on complex tasks, differentiation ceases or slows after the network is fitted to its task, and redundant layers generally fail to differentiate in the causal plane. Additionally, we show how the $EI$ can be used to calculate the difference between the total joint effects and the total individual effects of nodes in a layer, allowing for the measuring of feedfoward integrated information in a deep neural network \citep{oizumi2014phenomenology}.

The tools put forward here to assist in analyzing the causal structures of DNNs using information theory should assist with another central problem of the field, which is that large parameterizations often make DNNs into ``black boxes'' with millions of fine-tuned weights that allow for successful performance but that are impenetrable in their operations and functions \citep{gunning2017explainable}. A lack of explainability can mask other problems such as biases in either datasets \citep{alvi2018turning} or model choice \citep{mignan2019one}, and is a serious problem for those who want to use DNNs to make life and death decisions, such as in the case of self-driving cars \citep{bojarski2016end}, autonomous drones \citep{floreano2015science}, or medical diagnoses \citep{shin2016deep}. Using this suit of techniques, researchers will be able to directly observe the process during training wherein the overall causal structure of a DNN changes, a key step to opening up the ``black box'' and understanding what does what in DNNs.

\section{\label{sec:Quantifying_causal_structure}Quantifying the causal structure of DNNs}

\begin{figure}[t]
    \centering
    \includegraphics{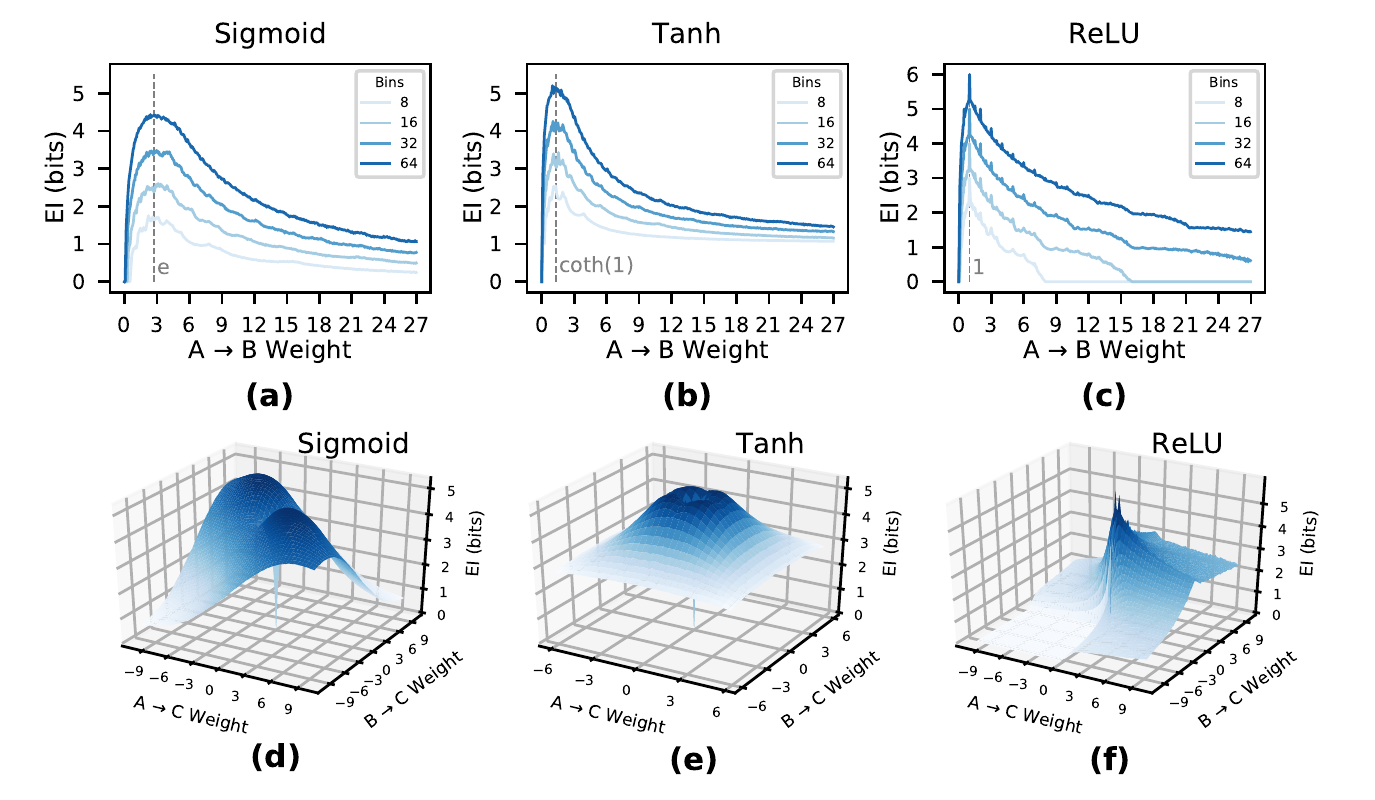}
    \caption{\textit{\textbf{$EI$ is a function of weights and connectivity}}. Plots \textbf{(a-c)} show $EI$ vs. weight for a single input and output neuron, using sigmoid, tanh, and ReLU activation functions, and computed using 8, 16, 32, and 64 bins. Marked is the most informative weights (in isolation) for transmitting a set of perturbations for each activation function. Plots \textbf{(d-f)} show $EI$ for a layer with two input nodes \textit{A}, \textit{B} and a single output nodes \textit{C}. Different activation functions have different characteristic $EI$ manifolds.}
    \label{Figure_1}
\end{figure}

Interventions (also called ``perturbations'') reveal causal relationships. The set of causal relationships (also called the``causal structure'') of a feedfoward DNN is composed of layers, their respective connections, and the activation functions of the nodes. We introduce tools to explore the hypothesis that the generalizability of DNNs is a matter of how their causal structures differentiate to fit the tasks they are trained on (all code is publicly-available, see \href{https://github.com/ei-research-group/deep-ei}{the repository here}).

To investigate this issue, we make use of a formal approach widely used to study causation where interventions are represented as the application of a $do(x)$ operator \citep{Pearl2000}. The $do(x)$ is normally used to set an individual variable in a given system, such as a directed acyclic graph, to a particular value (for instance, it has been used previously to apply individual interventions in DNNs \citep{harradon2018causal, narendra2018explaining}). Rather than tracking individual interventions, in order to generate an analytic understanding of the full causal structure of a DNN layer, we introduce here the use of an intervention distribution, $I_D$, which is a probability distribution over the $do(x)$ operator. The $I_D$ is simply a mathematical description of a set of interventions. The application of an $I_D$ over the inputs of a layer leads to some distribution of effects at the downstream outputs (the $E_D$) \citep{Hoel2017WhenTerritory}.

The informativeness of a causal relationship can be measured via information theory using an $I_D$. More informative causal relationships are stronger. Here, we make use of \textit{effective information} ($EI$), a measure of the informativeness of a causal relationship, to quantify and examine the causal structure of a layer. Specifically, the $EI$ is the mutual information between interventions and effects, $I(I_D,E_D)$, when $I_D=H^{\max}$, the maximum-entropy distribution. Put more simply, the $EI$ is the mutual information ($MI$) following a noise injection in the form of randomization.

Yet unlike the standard $MI$, which is a measure of correlation \citep{Shannon1948}, all mutual bits with a noise injection will necessarily be caused by that noise. Additionally, as the maximally-informative intervention (in terms of its entropy), $EI$ represents the information resulting from the randomization of a variable, which is the gold standard for causation in the sciences \citep{Fisher1936}. It can be thought of as measuring how well the image of the function can be used to recover the pre-image, and has important relationships to Kolmogorov Complexity and VC-entropy \citep{balduzzi2011information}. Most notably, previous research has shown that $EI$ reflects important properties for causal relationships, capturing how informative a causal relationship is, such as their determinism (lack of noise) or degeneracy (lack of uniqueness) \citep{hoel2013quantifying}.

First, we introduce a way to measure the $EI$ of layer-to-layer connectivity in a DNN, capturing the total joint effects of one layer on another. Therefore we start with \textit{$L_1$}, which is a set of nodes that have some weighted feedforward connection to \textit{$L_2$}, and we assume that all nodes have some activation function such as a sigmoid function. In order to measure $EI$, \textit{$L_1$} is perturbed at maximum entropy, $do(L_1=H^{\max})$, meaning that all the activations of the nodes are forced into randomly chosen states. $L_1=H^{\max}$ implies simultaneous and independent maximum-entropy perturbations for all nodes $i$ in $L_1$:

\begin{equation}\label{eq:ei}
    EI = I(L_1,L_2) \ | \ do(L_1=H^{\max})
\end{equation}

That is, the calculation is made by measuring the mutual information between the joint states of $L_1$ and $L_2$ under conditions of $L_1=H^{\max}$.

$EI$ scales across different commonly-used activation functions. Fig. \ref{Figure_1}a-c shows the $EI$ of a single edge between two nodes, $A$ and $B$, wherein $A \rightarrow B$ with increasing weight, with each panel showing a different activation function (sigmoid, tanh, ReLU). We can see that for each isolated edge with a given activation function there exists a characteristic $EI$ curve dependent on the weight of the connection from \textit{A} to \textit{B}, and that the shape of this curve is independent of the number of bins chosen (8, 16, 32, and 64). At low weights, the $EI$ shows that \textit{B} is not sensitive to perturbations in \textit{A}, although this sensitivity rises to a peak in all three activation functions. The curve then decays as the weight saturates the activation function, making \textit{B} insensitive to perturbations of \textit{A}.

Note that the characteristic peaks reveal which weights represent strong causal relationships (of a connection considered in isolation). For instance, a sigmoid activation function has the most informative causal relationship at a weight equal to Euler's number $e$, a tanh activation function at weight $\coth(1)$, a ReLU activation function at weight 1. This indicates the most important weights in a DNN may be the most causally efficacious, not the highest in absolute value. For example, with sigmoid activation functions and an extremely high weight connecting $A \rightarrow B$, $A$'s activation is not very informative to perturb, since most perturbations will lead to a saturation of $B$'s output at 1. 

In the case of multiple connections the $EI$ curve becomes a higher-dimensional $EI$ manifold. Fig. \ref{Figure_1}d-f shows the $EI(A,B \rightarrow C)$ of a layer comprised of two nodes ($A$, $B$) each with a single connection to $C$. Since perturbations can interfere with one another, the $EI$ depends not only on the sensitivity of the relationships between nodes, but also the overlap, or $degeneracy$, of the network connectivity, thus creating a manifold. For instance, in sigmoid activation functions, the $EI$ manifold is roughly 2-fold symmetric, which is due to the symmetric nature of the sigmoid around positive and negative weights, combined with the symmetric nature of the network itself, as both neuron \textit{A} and \textit{B} only connect to \textit{C}.

Note that while the number of bins determines the amplitude of the curve, the rise / decay behavior is consistent across them, indicating that as long as bin size is fixed at some chosen value, ratios and behavior will be preserved (Figure \ref{Figure_1} uses 30,000 timesteps for the noise injection for a-c and 100,000 samples for d-f). $EI$ values for a DNN layer tend to converge to a particular value if the noise injection is long enough and the bin size is high enough. Evidence for this can be found in SI Section 1.

First however, we assess how changes to $EI$ occur during training networks on common machine learning tasks.

\section{\label{sec:EI_training}Information in the causal structure changes during training}

\begin{figure}[t]
    \centering
    \includegraphics{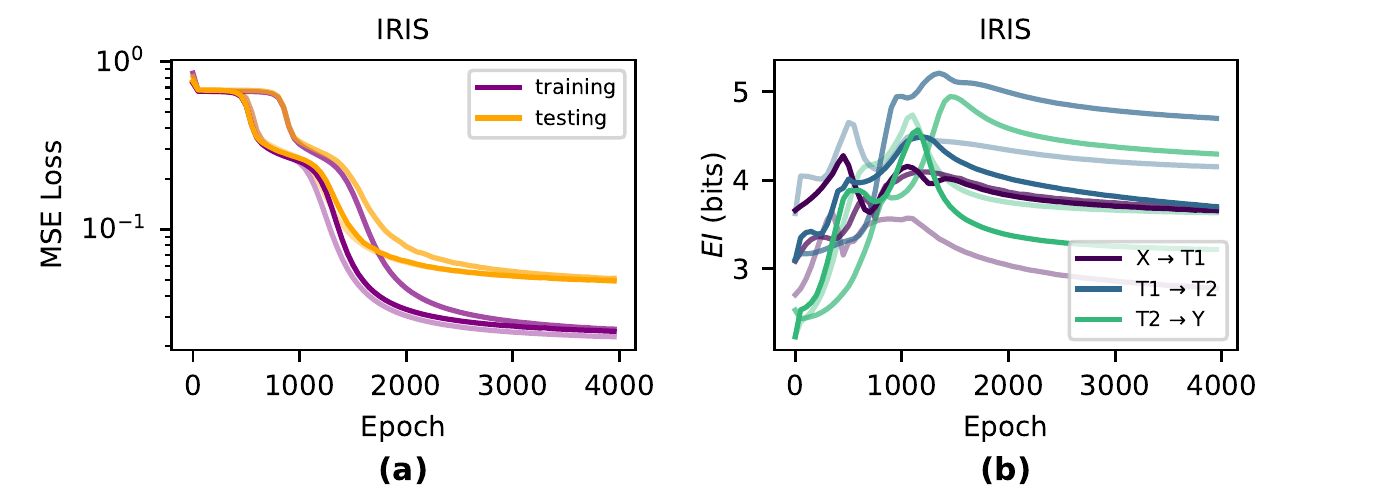}
    \includegraphics{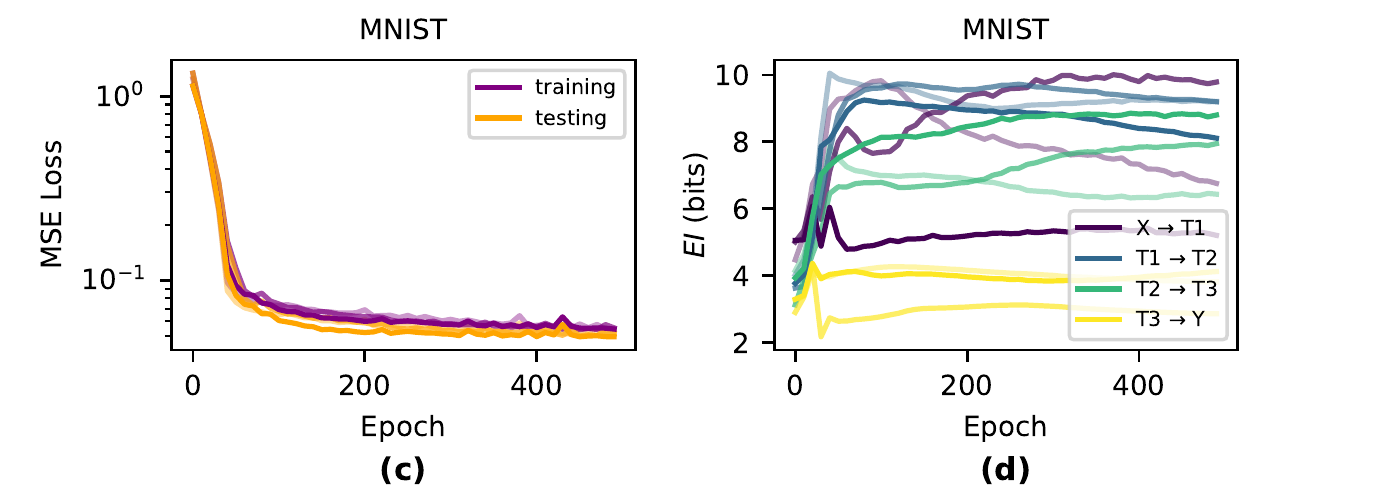}
    \caption{\textit{\textbf{How $EI$ and evolves during training across three different runs}.} Notably the largest changes in $EI$ occur during the steepest reductions in the loss function for both Iris-trained networks (top) and MNIST-trained networks (bottom).}
    \label{Figure_2}
\end{figure}

To understand how the causal structures of DNNs change during learning, we tracked the $EI$ in networks trained on two benchmark classification tasks: Iris \citep{Fisher1936} and MNIST \cite{lecun2010mnist}. For Iris, we trained networks with three densely connected layers $4 \rightarrow 5 \rightarrow 5 \rightarrow 3$ and for MNIST we used networks with four densely connected layers $25 \rightarrow 6 \rightarrow 6 \rightarrow 5$, using sigmoid activation functions and no biases for both tasks. For MNIST, we reshaped the inputs from 28x28 down to 5x5, and removed examples of digits 5-9 from the dataset so that the final layer has only 5 nodes -- this was necessary in order to reduce the computational cost of accurately computing $EI$. Networks for both tasks were trained with MSE loss and vanilla gradient descent with a learning rate of $0.01$. We trained the Iris networks with a batch-size of 10 for 4000 epochs and the MNIST networks with a batch-size of 50 for 500 epochs. We initialized the weights by sampling from the uniform distribution $W_{ij} = \mathcal{U}([-\frac{1}{\sqrt{\mathrm{fan}_{\mathrm{in}}}}, \frac{1}{\sqrt{\mathrm{fan}_{\mathrm{in}}}}] )$. For each task and architecture, we perform three runs with distinct initializations. Using the same respective network architectures, we also trained networks with tanh and ReLU activation functions -- results can be found in SI Section \ref{sec:EI_activations}. To compute $EI$, we use a fixed noise injection length of $10^7$ samples. We found that in our networks, an injection of this length was enough to ensure convergence (see SI Section \ref{sec:convergence_SI}). Note, however, that wider network layers may require many more samples. 

Qualitatively, we observe that the greatest changes in $EI$ significantly match the steepest parts of the loss curve during training and $EI$ is generally dynamic during periods of greatest learning (shown in Figure \ref{Figure_2}). During the overfitting period when training performance dissociated from testing performance, $EI$ was generally flat across all layers, indicating that the information in the causal structure was unchanged during this period after the network had appropriately fitted.

\section{\label{sec:sensitivity_degeneracy}Deep neural networks in the causal plane}

\begin{figure}[t]
    \centering
    \includegraphics{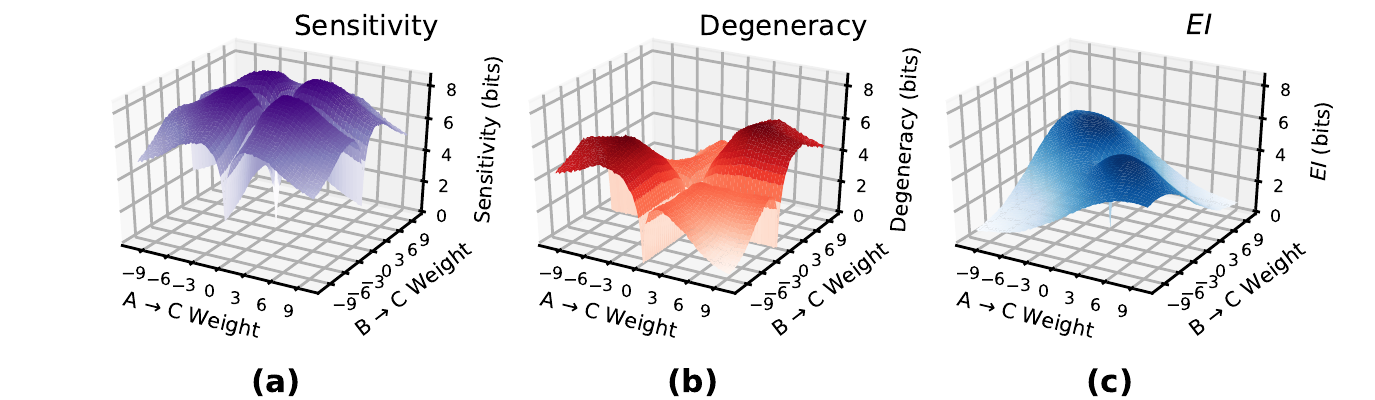}
    \caption{\textbf{\textit{EI is composed by sensitivity and degeneracy}}. The above surfaces are the sensitivity and degeneracy of a layer with two input nodes and a single output nodes, with a sigmoid activation function. Subtracting the surface \textbf{(b)} from the surface \textbf{(a)} gives the $EI$ manifold as in \textbf{(c)}}
    \label{Figure_4}
\end{figure}

As discussed in Section \ref{sec:Quantifying_causal_structure}, $EI$ depends both on the weight of connections as well as their degree of overlap, which together create the $EI$ manifold. This indicates that $EI$ can be decomposed into two properties: the $sensitivity$ of the causal relationships represented by individual weights and the $degeneracy$ of those relationships due to overlap in input weights. This mirrors previous decompositions of the $EI$ in Boolean networks or Markov chains into the determinism (here replaced with sensitivity, since neural networks are traditionally deterministic) and degeneracy \citep{hoel2013quantifying,klein2020emergence}.

In DNNs, the $sensitivity$ of a layer measures how well the input transmits perturbations to the output nodes, while the $degeneracy$  of a layer measures how well the source of input perturbations can be reconstructed by examining the layer output. If the source of a perturbation cannot be reconstructed well the network is said to be \textit{degenerate}. Together, these two dimensions of causal relationships form a ``causal plane'' which all DNN layers occupy. As layers differentiate via learning, their causal structures should occupy unique positions in the causal plane reflecting their contribution to the function of the DNN by becoming more sensitive or more degenerate. 

To identify the position or trajectory of a DNN layer in the causal plane, both $sensitivity$ and $degeneracy$ are explicitly calculated based on the components of $EI$. The $sensitivity$ is calculated by summing the total contribution of each edge individually, in the absence of interaction effects between parameters. Therefore, the total $sensitivity$ from layer $L_1$ to the next layer $L_2$ is:

\begin{equation}\label{eq:sensitivity}
    Sensitivity = \sum_{(i \in L_1, j \in L_2)} I(t_i,t_j) \ | \ do(i=H^{\max})
\end{equation}

This is the same as calculating the $EI$ of each ($i$,$j$) pair, but done independently from the rest of the network. Note that in a layer wherein each node receives only one unique input (i.e., no overlap) the $sensitivity$ is equal to the $EI$.

The $degeneracy$ of a layer measures how much information in the causal relationships is lost from overlapping connections, and is calculated algebraically as $sensitivity - EI$, since $sensitivity$ measures the information contribution from non-overlapping connections in the network. Figure \ref{Figure_4} shows $sensitivity$ and $degeneracy$ manifolds for a layer of two input nodes and one output node (with sigmoid activations) with varying connection weights. The difference between them creates the $EI$ manifold.

Previous research investigating the $EI$ of graphs (based on random walk dynamics) has led to a way to classify different canonical networks, such as Erd\H{o}s-R\'enyi random graphs, scale-free networks, and hub-and-spoke models, based on where they fall in terms of the determinism and degeneracy of random walkers \citep{klein2020emergence}. For $EI$ in DNNs a $sensitivity$ term takes the place of determinism.

In order to visualize layer shifts between $sensitivity$ and $degeneracy$ we introduce the ``causal plane'' of a DNN wherein the two dimensions of the plane represent the two respective values. The causal plane makes use of the fact that, since $EI = sensitivity - degeneracy$, if both increase equally, the $EI$ itself is unchanged. When $degeneracy$ vs. $sensitivity$ is plotted, points on the line $y = x$ represent zero $EI$, and we refer to this $45^{\circ}$ line as the ``nullcline'' of the $EI$. Paths that move more towards sensitivity will increase $EI$, and paths that move more towards degeneracy will decrease $EI$, while paths along the $EI$ nullcline will not change $EI$.

\begin{figure}[t]
    \centering
    \includegraphics{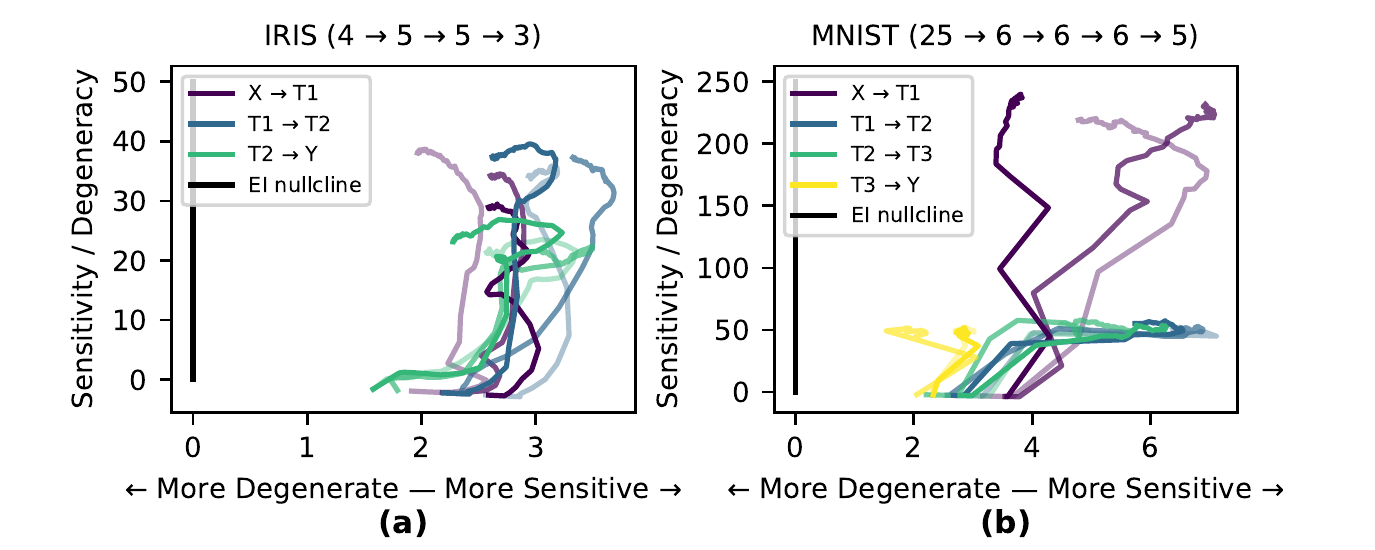}
    \includegraphics{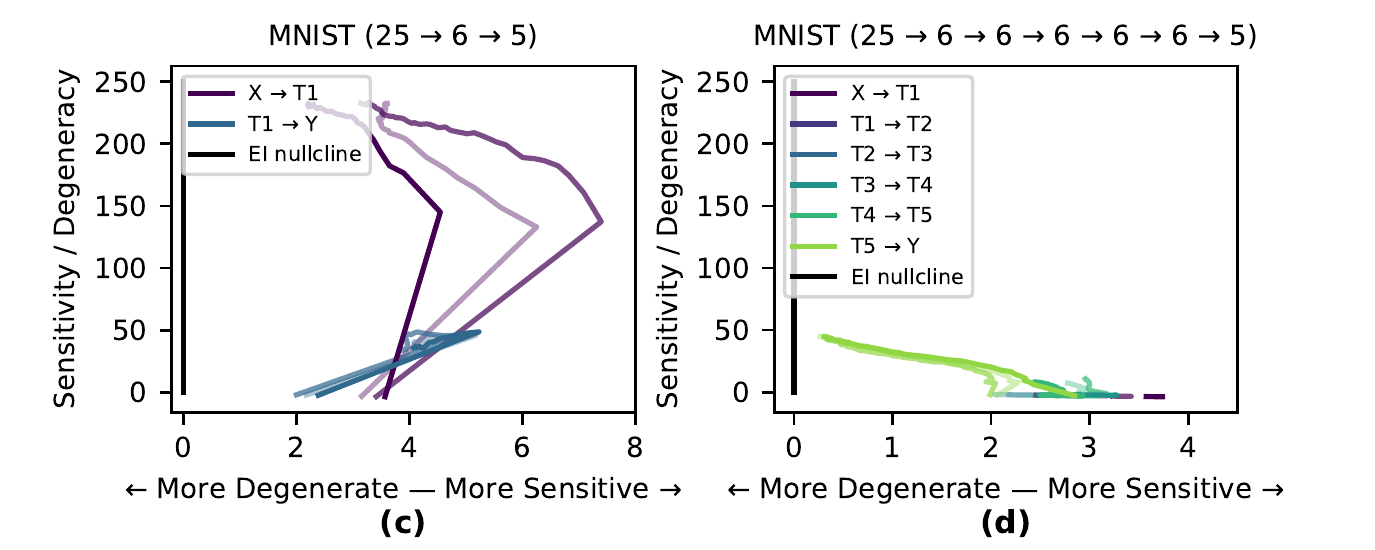}
    \caption{\textbf{\textit{Behavior on the causal plane during training}.} Paths traced on the causal plane in different layers. All paths get less smooth over time during the period of overfitting and move about less in the causal plane. Networks trained on the simpler Iris task show less differentiation between layers than those trained on the MNIST task.}
    \label{fig:causal_plane}
\end{figure}

Here we explore the hypothesis that the internal causal structure of a DNN shifts to match the task it is trained on, and that this happens in specific stages throughout the training process. To investigate this, we measured the paths of three runs on the Iris and MNIST data sets through the causal plane during training (shown in Fig. \ref{fig:causal_plane}a-b). Of the two tasks, classifying MNIST digits is more degenerate and complex, as the network must transform a manifold in a high dimensional space into only 10 distinct output classes (or rather 5 for our reduced version of MNIST here). The task of classifying Iris flowers is not as degenerate nor complex, as the network must transform a 4 dimensional space into 3 (mostly) linearly separable classes. If a network learns by matching its internal causal structure to the data set a network trained on MNIST would shape itself to a greater degree than one trained on Iris. This is precisely what we observe in Figure \ref{fig:causal_plane} wherein the MNIST-trained network shows much greater differentiation and movement within the causal plane, while there is less differentiation in the causal structure of the Iris-trained network as it follows the $EI$ nullcline. In many cases, particularly for hidden and output layers, the runs first demonstrate an increase in sensitivity (increasing the $EI$), and then later an increase in degeneracy (decreasing the $EI$).

In order to examine the hypothesis that the causal structure of layers necessarily differentiate in response to training, the MNIST-trained network with sigmoid activation functions was modified in two ways: in one case a hidden layer was removed, and in the other case a number of redundant hidden layers were added (Fig. \ref{fig:causal_plane}c-d). Both modifications of the network trained as accurately as the previous network. In the causal plane the added redundant layers moved very little, indicating a net-zero contribution to the $EI$ during training (for movie see \href{https://github.com/ei-research-group/deep-ei}{the GitHub}). This shows how redundant layers that don't contribute to the network's causal structure cluster along the $EI$ nullcline and move little, compared to more dynamic layers.

\section{\label{sec:integrated_information}Measuring joint effects of layer-to-layer connectivity}

Integrated Information Theory (IIT) has been used to assess the total information contained in joint effects versus their independent effects in systems \citep{tononi2008consciousness}. It is a useful tool for causal analysis, analyzing the amount of information being integrated in a network's causal structure \citep{marshall2017causal, albantakis2019caused}. Previously, the integrated information has been measured as the loss in $EI$ given a partition \citep{balduzzi2008integrated}, making $EI$ the upper bound for integrated information. However, there is no one accepted and universal measure of integrated information \citep{oizumi2014phenomenology, oizumi2016unified}. Instead, various measures for integrated information have been put forward in different systems \citep{tegmark2016improved, mediano2019measuring}. Traditionally the amount of integrated information in a feedfoward network is zero since there is no reentrant connectivity, since it is based on finding the minimum information partition across all possible subsets of a system. However, even in a feedforward network a layer's nodes can still contain irreducible joint effects on another layer, and therefore we introduce a measure, feedfoward integrated information, to apply in DNNs.

Normally calculating the integrated information requires examining the set of all possible partitions, which prohibits this method for systems above a small number of dimensions. Alternatively, in order to assess the synergistic contribution to $EI$ of individual edges, one would likely need to use multi-variate information theory, such as the partial information decomposition, which grows at the sequence of Dedekind numbers as sources are included \citep{williams2010nonnegative}. 

In order to avoid these issues we introduce a measure, $EI_{parts}$, which is calculated based on contributions of each edge. That is, for each node $i \in L_1$ the time-series $t_i$ of its activation function under this perturbation is recorded, along with that of each node $j \in L_2$. To calculate $EI_{parts}$, each individual time-series of each node is then discretized into some shared chosen bin size, and the $MI$ of each ($i$,$j$) pair is calculated and summed:

 \begin{equation}\label{eq:timeseries}
 EI_{parts}(L_1 \rightarrow L_2) = \sum_{(i \in L_1,j \in L_2)} I(t_i,t_j) \ | \ do(L_1=H^{\max}).
\end{equation}

Note that for a layer with a single node, $EI$ and $EI_{parts}$ are identical. The same is true when each node of the network only receives a single edge. However, $EI_{parts}$ measure will necessarily miss certain positive joint effects. Importantly, the difference between $EI$ and $EI_{parts}$ measures can capture the amount of joint effects and therefore the amount of information the layer-to-layer is integrating in a feedforward manner. Specifically, we compare $EI$, the upper bound for integrated information, to $EI_{parts}$ as defined in Section \ref{sec:EI_training}, that is $\phi_{feedforward}=EI - EI_{parts}$. It should be noted that $\phi_{feedforward}$, while designed to capture total joint effects of one layer to another, is not bounded by zero and can be negative.

\begin{figure}[t]
    \centering
    \includegraphics{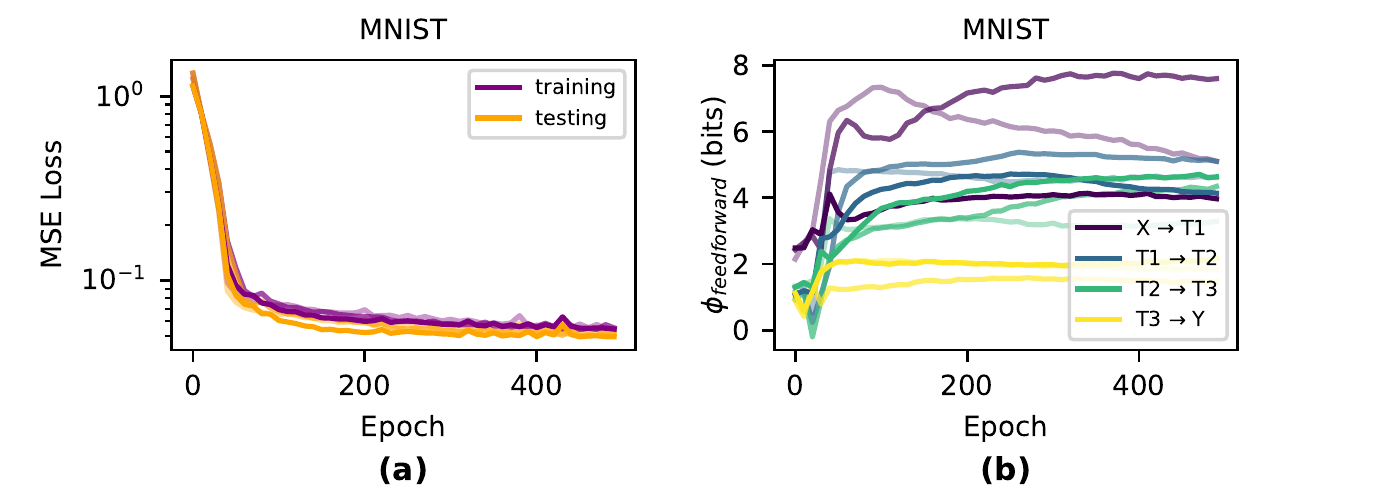}
    \includegraphics{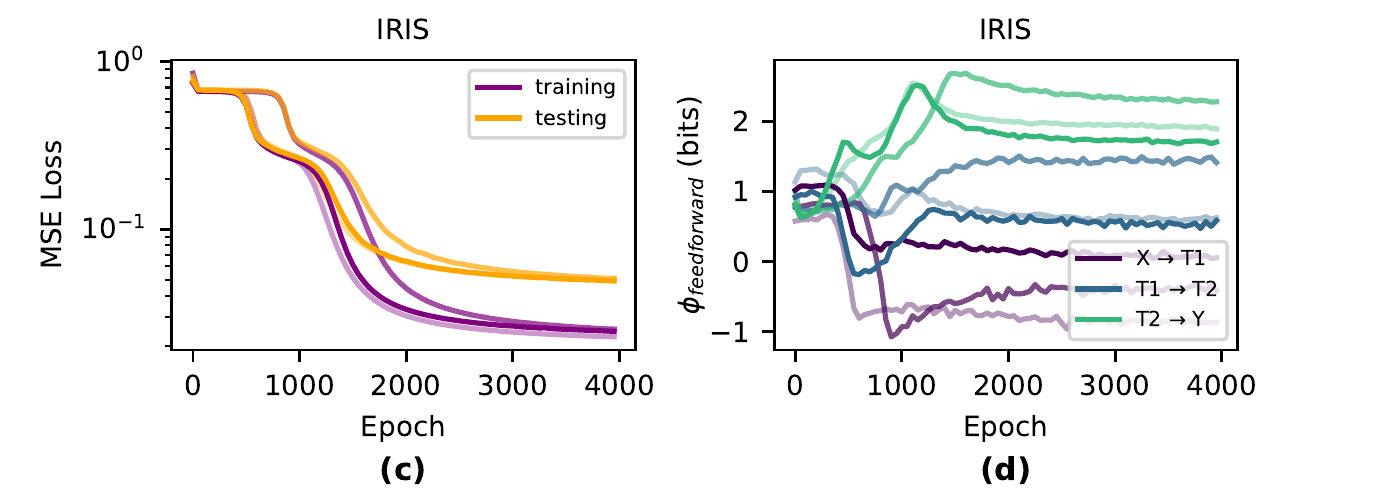}
    \caption{\textbf{\textit{Integrated Information over training.}} MNIST-trained networks develop more $\phi_{feedforward}$ during training than IRIS-trained networks.}
    \label{Figure_6}
\end{figure}

To understand how layer-to-layer joint effects change during training of a DNN, we analyzed how $\phi_{feedforward}$ changes during training across both Iris and MNIST data sets (see SI Section \ref{sec:convergence_SI} for details on our methodology for measuring $EI_{parts}$). We observe that MNIST-trained networks have higher $\phi_{feedforward}$ than Iris-trained networks, indicating that the causal structure has indeed differentiated in accordance with the complexity of the task and requires more joint effects to learn (Figure \ref{Figure_6}).

\section{\label{sec:discussion}Discussion}

Here we have introduced information-theoretic techniques to categorize and quantify the causal structures of DNNs based on information flows following perturbations. These techniques are built around the effective information ($EI$), which we adapted to apply to DNNs. It is defined as the mutual information following a set of perturbations of maximum entropy, and it reveals the information contained in the causal structure of a layer. For networks trained on both Iris and MNIST tasks, $EI$ changed during the training period, particularly when learning actually occurred (as reflected by step changes in the loss function).

$EI$ depends on both the $sensitivity$ and $degeneracy$ of a network. The $sensitivity$ between two nodes reflects the strength of causal relationships in isolation, and peaks at particular characteristic weights for different activation functions (e.g., in sigmoid activation functions it peaks at $e$). The $degeneracy$ of a layer reflects the difficulty of downstream reconstruction of an upstream perturbation due to overlap of edge weights. Analyzing the $EI$ reveals where networks lie on sensitivity/degeneracy space, which we call the ``causal plane.'' The ability to place network architectures in this plane means we can track how any given DNN’s causal structure evolves during its training as it moves through the space. Our results indicate that the causal structure of an DNN reflects the task it is trained on. For instance, in the MNIST task, different layers have a clear task in the causal structure of the DNN, reflected by each layer's different trajectory in the causal plane, and adding new redundant layers added no new information to the causal structure by not contributing to the $EI$. 

These techniques offer a different approach than work on information bottlenecks \citep{tishby2015deep}, which is focused on using the mutual information to measure correlations between inputs and node activity. Both approaches have a similar goal to explain DNN generalizability and both share formal similarities, although here the focus is on the layer-by-layer causal structure itself rather than the input of DNNs. In the future this work can be extended to different activation functions beyond the three considered here \citep{karlik2011performance, nair2010rectified}, unsupervised tasks \citep{wiskott2002slow}, recurrent neural networks such as LSTMs \citep{hochreiter1997long}, and convolutional neural networks \citep{krizhevsky2012imagenet}. 

These techniques open up the possibility of assessing decompositions and expansions of the $EI$, such as the integrated information of DNNs (since integrated information can be calculated using the minimum of $EI$ between subsets of a network \citep{tononi2003measuring}), and integrated information is also decomposable into properties similar to $sensitivity$ and $degeneracy$ \citep{hoel2016can}. Here, a measure of integrated information, $\phi_{feedforward}$, is outlined that measures the irreducible joint effects in feedforward layer connectivity. 

All of these may help understand why certain network architectures generalize and why some do not. In the future these techniques also open the possibility for direct measurement of individual instances of causation in DNNs \citep{albantakis2019caused}.

\section*{Acknowledgements}
\textbf{Funding:} This publication was made possible through the support of a grant from the Army Research Office (proposal 77111-PH-II). This research was also supported by the Allen Discovery Center program through The Paul G. Allen Frontiers Group (12171). \textbf{Author contributions:} S.M., E.J.M., and E.H. conceived the ideas and wrote the article. S.M. and E.J.M. created the code. E.J.M. made the figures and did the analyses. \textbf{Competing interests:} The authors declare no competing interests.

\bibliographystyle{plainnat}
\bibliography{biblio}

\newpage

\section{\label{sec:SI}Supplementary information}
\subsection{\label{sec:convergence_SI}Effective information converges across measurement schemes and can be found via extrapolation}

\begin{figure}[h]
    \centering
    \includegraphics[width=0.5\textwidth]{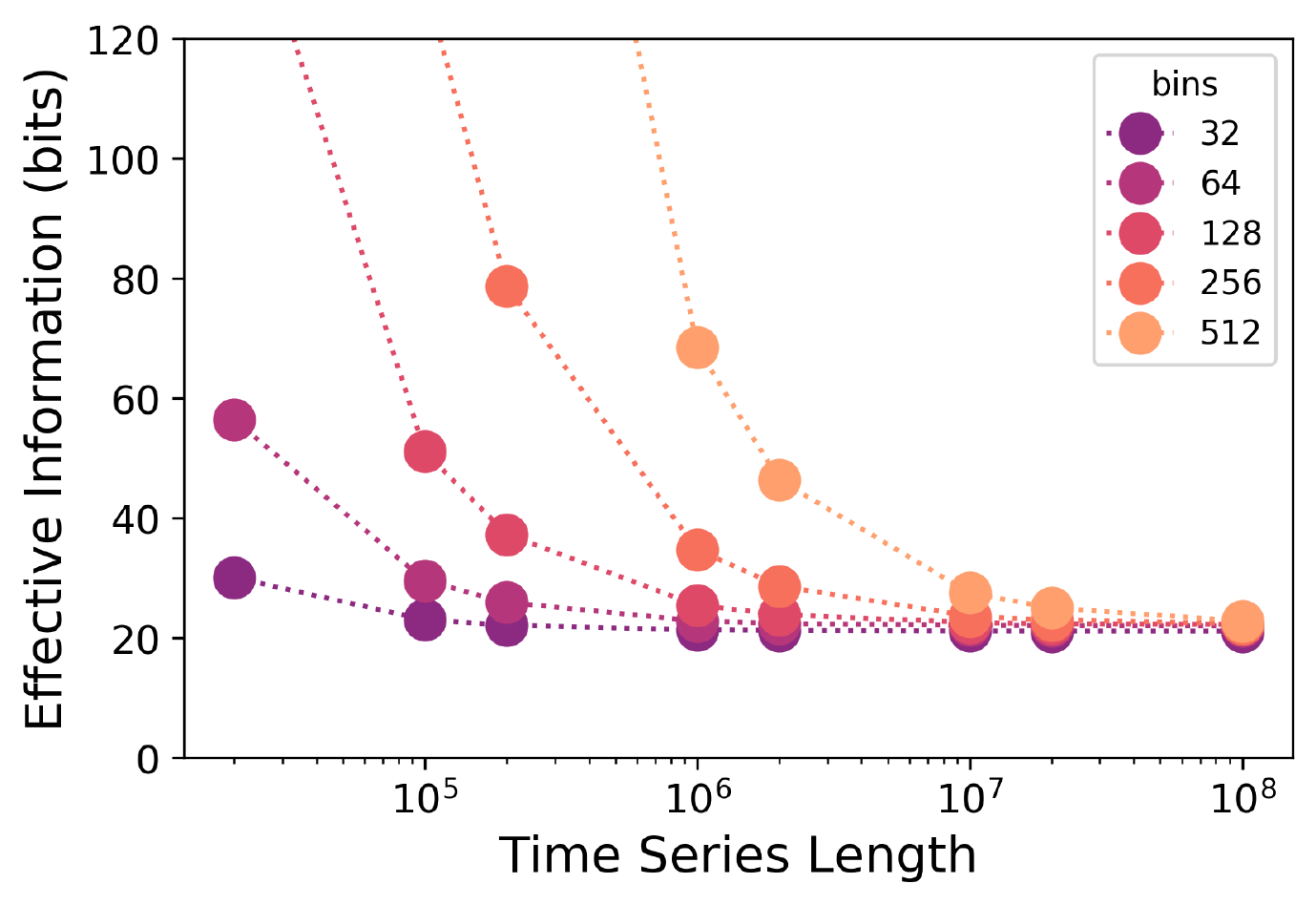}
    \caption{\textbf{Convergence of $EI_{parts}$ measures to theoretical values.} The $EI_{parts}$ a $30 \rightarrow 30$ layer injected with a time-series of noise up to $10^8$ time-steps and analyzed with different numbers of bins.}
    \label{fig:SI_convergencevaryingbins}
\end{figure}

Here $EI$, $EI_{parts}$, and \emph{sensitivity} are calculated based off of an injection of noise into a layer. This requires a choice of both the amount of time spent randomizing the input (the number of noise samples used), as well as the binning scheme of all nodes. In Figure \ref{fig:SI_convergencevaryingbins}, we examined how $EI_{parts}$ converges for a $30 \rightarrow 30$ dense layer with varying number of bins. The layer was initialized with the uniform distribution from earlier (Section \ref{sec:EI_training}). As we see in Figure \ref{fig:SI_convergencevaryingbins}, provided enough bins are used, $EI_{parts}$ generally converges to about the same value regardless of the exact number of bins used. However, the number of noise samples which must be injected for the $EI_{parts}$ to converge greatly increases with the number of bins. With 256 bins convergence of $EI_{parts}$ can sometimes take millions of samples, and one must therefore be careful about specifying a precise number of samples to use when computing $EI_{parts}$. 

\begin{figure}
    \centering
    \includegraphics{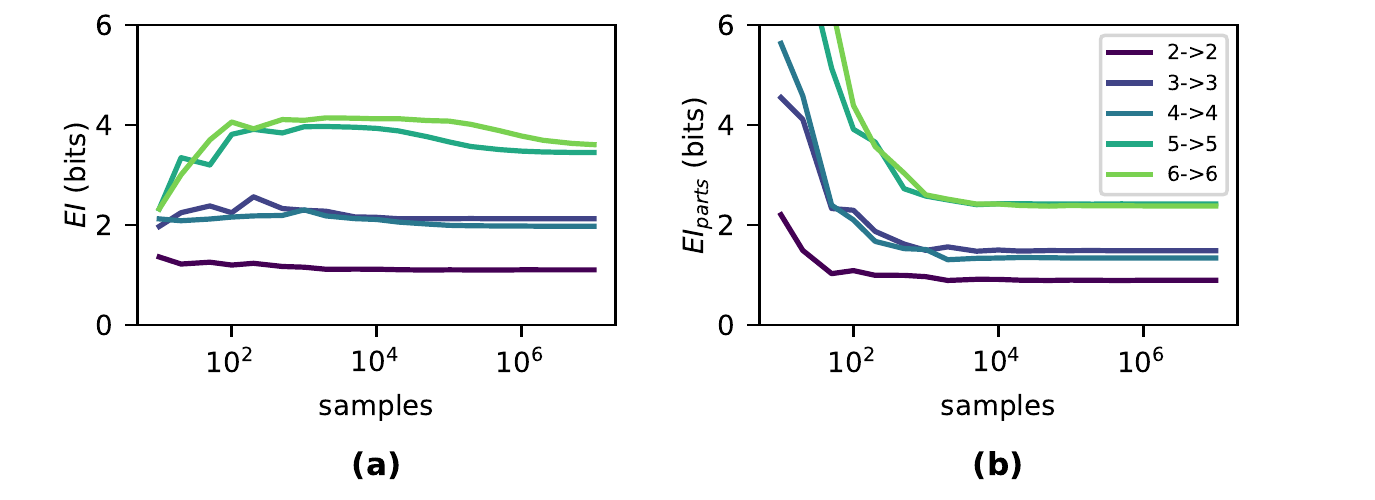}
    \includegraphics{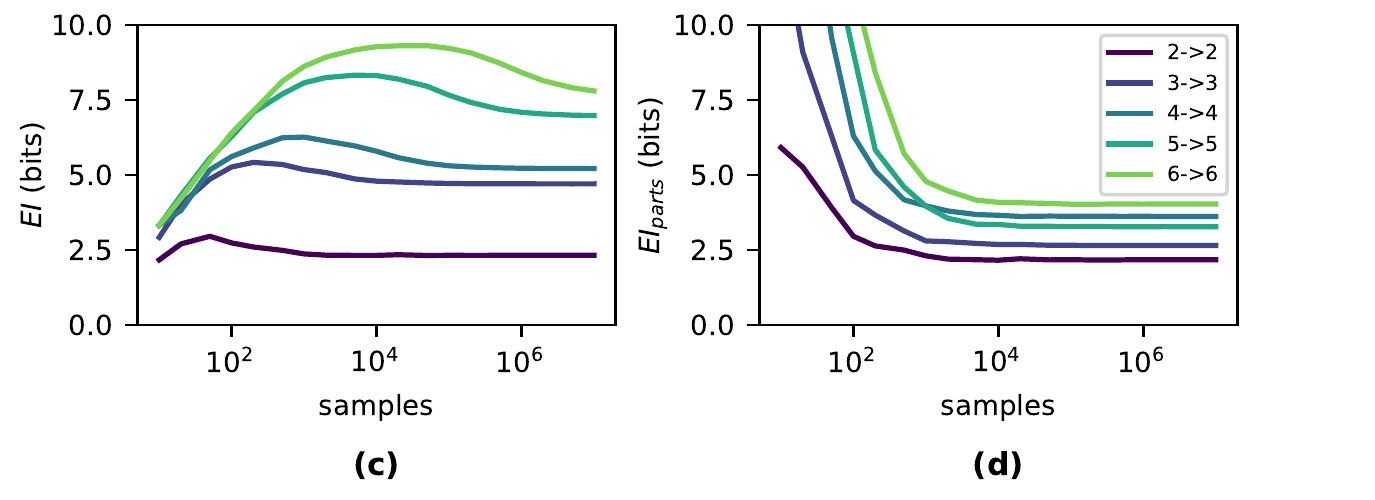}
    \caption{\textbf{Convergence of $EI$ and $EI_{parts}$.} If evaluated on enough noise samples, $EI$ and $EI_{parts}$ converge. In \textbf{(a)} and \textbf{(b)}, we show how $EI$ and $EI_{parts}$ respectively converge for dense layers of varying width, initialized with the distribution $\mathcal{U}([-\frac{1}{\sqrt{\mathrm{fan}_{\mathrm{in}}}}, \frac{1}{\sqrt{\mathrm{fan}_{\mathrm{in}}}}])$. In figures \textbf{(c)} and \textbf{(d)}, we show the same, but with weights sampled from  $\mathcal{U}([-\frac{5}{\sqrt{\mathrm{fan}_{\mathrm{in}}}}, \frac{5}{\sqrt{\mathrm{fan}_{\mathrm{in}}}}])$. 
    }  
    \label{fig:SI_convergence}
\end{figure}

To accurately compute $EI_{parts}$ without having to specify a fixed number of samples, we used two techniques. When it was computationally tractable (which it was for all the experiments presented here), we successively double the number of samples used in the injection until the expected change (computed with secant lines through the $EI_{parts}$ vs. samples plot) in $EI_{parts}$ of another doubling is less than 5\% of the most recently-computed value. In some scenarios, this technique, which computes $EI$ directly, requires many millions of samples (or as many as are needed for the $EI_{parts}$ vs. samples line to level off) and therefore is often intractable for large densely-connected layers, or when a large number of bins are used. As a more tractable alternative, for the larger layers (like those in our MNIST-trained networks) we introduced a way to measure $EI_{parts}$ with varying numbers of samples and fit a curve to the $EI_{parts}$ vs samples relationship. Across a range of layer connectivities and sizes, we observe that the $EI_{parts}$ vs. samples curve takes the form:
$$ EI_{parts}(s) = \frac{A}{s^\alpha} + C$$
To extrapolate $EI_{parts}$, we evaluate $EI_{parts}$ directly on $100$K, $200$K, \ldots, $2M$ samples, then fit the above curve, and evaluate it at $10^{15}$. While this method does not compute $EI_{parts}$ directly, we find that in practice it gives accurate values. 

Note that these methods apply only to the computation of $EI_{parts}$ which we find to be monotonically decreasing in the number of samples used to compute it. Computing the full $EI$ is in general a much harder problem. Figure \ref{fig:SI_convergence} shows convergence curves for both $EI$ and $EI_{parts}$ for layers of varying width, computed with 8 bins per node. As the number of samples used increases, $EI$ at first increases before decreasing and leveling off by $10^7$ samples in layers of width no greater than 6 neurons.

\subsection{\label{sec:EI_activations}Effective information tracks changes in causal structure regardless of activation function}

\begin{figure}[h]
    \centering
    \includegraphics{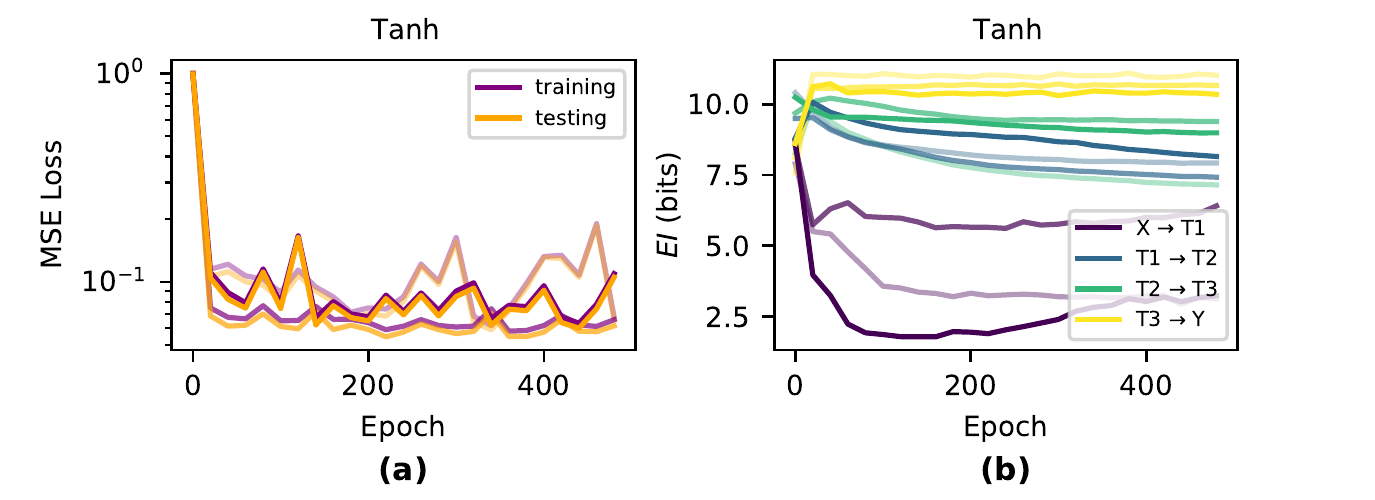}
    \includegraphics{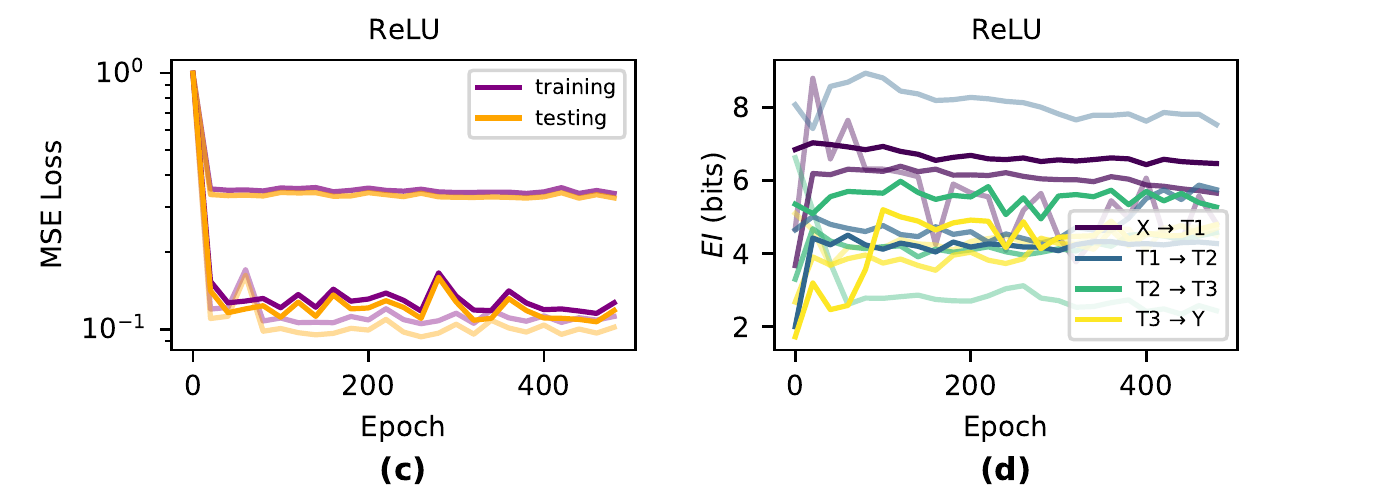}
    \caption{\textbf{Changes in $EI$ during training across activation functions.} Tanh (top) and ReLU (bottom) versions of a network trained on the reduced-MNIST task.}
    \label{fig:different_activations_eiwhole}
\end{figure}

Causal relationships should depend on activation functions. To test this, we further examined the $EI$ of Iris and MNIST-trained networks, yet with tanh and ReLU activation functions (shown in Figure \ref{fig:different_activations_eiwhole}). Despite using different initializations, training order, and activation functions, the changes in $EI$ during training were broadly similar, although each choice of activation function changed precise behavior in $EI$.

\end{document}